\documentclass[sigconf]{acmart}
\usepackage{graphicx}
\usepackage{amsmath}
\usepackage{booktabs}
\usepackage{arydshln}
\usepackage{bbold}
\usepackage{algorithmicx,algorithm}
\usepackage{algorithm}
\usepackage{algpseudocode}
\usepackage{colortbl}
\usepackage{multirow}
\usepackage{xcolor}
\usepackage{array}
\usepackage{float}
\usepackage{caption}
\usepackage{subcaption}
\usepackage{url}

\definecolor{mygray}{rgb}{.929, .929, .913}
\newcommand{\CC}{\cellcolor{mygray}}
\newcommand{\eg}{e.g., }
\newcommand{\ie}{i.e., }
\AtBeginDocument{%
}

\copyrightyear{2023}
\acmYear{2023}
\setcopyright{acmlicensed}
\acmConference[MM '23]{Proceedings of the 31st ACM International Conference on Multimedia}{October 29-November 3, 2023}{Ottawa, ON, Canada}
\acmBooktitle{Proceedings of the 31st ACM International Conference on Multimedia (MM '23), October 29-November 3, 2023, Ottawa, ON, Canada}
\acmPrice{15.00}
\acmDOI{10.1145/3581783.3611713}
\acmISBN{979-8-4007-0108-5/23/10}
\acmSubmissionID{99}

\begin{document}

\title{CUCL: Codebook for Unsupervised Continual Learning}


\author{Cheng Chen}
\affiliation{%
  \institution{Center for Future Media, University of Electronic Science and Technology of China}
  \city{Chengdu}
  \country{China}
}
\email{cczacks@gmail.com}

\author{Jingkuan Song}
\authornote{Corresponding Author.}
\affiliation{%
  \institution{Center for Future Media, University of Electronic Science and Technology of China}
  \city{Chengdu}
  \country{China}
}
\email{jingkuan.song@gmail.com}

\author{Xiaosu Zhu}
\affiliation{%
  \institution{Center for Future Media, University of Electronic Science and Technology of China}
  \city{Chengdu}
  \country{China}
}
\email{xiaosu.zhu@outlook.com}

\author{Junchen Zhu}
\affiliation{%
  \institution{Center for Future Media, University of Electronic Science and Technology of China}
  \city{Chengdu}
  \country{China}
}
\email{junchen.zhu@hotmail.com}

\author{Lianli Gao}
\affiliation{%
  \institution{Shenzhen Institute for Advanced Study, University of Electronic Science and Technology of China}
  \city{Shenzhen}
  \country{China}
}
\email{lianli.gao@uestc.edu.cn}

\author{Hengtao Shen}
\affiliation{%
  \institution{Shenzhen Institute for Advanced Study, University of Electronic Science and Technology of China}
  \city{Shenzhen}
  \country{China}
}
\email{shenhengtao@hotmail.com}
\renewcommand{\shortauthors}{Cheng Chen et al.}
%%\renewcommand{\shortauthors}{Trovato et al.}

\begin{abstract}
The focus of this study is on Unsupervised Continual Learning (UCL), as it presents an alternative to Supervised Continual Learning which needs
high-quality manual labeled data.
The experiments under UCL paradigm indicate a phenomenon where the results on the first few tasks are suboptimal. 
This phenomenon can render the model inappropriate for practical applications. 
To address this issue, after analyzing the phenomenon and identifying the lack of diversity as a vital factor, we propose a method named \textbf{Codebook for Unsupervised Continual Learning (CUCL)} which promotes the model to learn discriminative features to complete the class boundary.
Specifically, we first introduce a \textbf{Product Quantization} to inject diversity into the representation and apply a cross quantized contrastive loss between the original representation and the quantized one to capture discriminative information. 
Then, based on the quantizer, we propose a effective \textbf{Codebook Rehearsal} to address catastrophic forgetting.
% In addition, we propose a new metric - Mean Average Accuracy (MAA), which measures performance precisely throughout the training process.
This study involves conducting extensive experiments on CIFAR100, TinyImageNet, and MiniImageNet benchmark datasets. 
Our method \textbf{significantly} boosts the performances of supervised and unsupervised methods.
For instance, on TinyImageNet, our method led to a relative improvement of 12.76\% and 7\% when compared with Simsiam and BYOL, respectively.
Codes are publicly available at \url{https://github.com/zackschen/CUCL}.
\end{abstract}

\begin{CCSXML}
<ccs2012>
<concept>
<concept_id>10010147.10010178.10010224.10010240.10010241</concept_id>
<concept_desc>Computing methodologies~Image representations</concept_desc>
<concept_significance>500</concept_significance>
</concept>
</ccs2012>
\end{CCSXML}

\ccsdesc[500]{Computing methodologies~Image representations}

\keywords{Continual learning, Unsupervised learning, Lifelong learning, Point quantization, Contrastive learning.}


\maketitle

\section{Introduction}
\label{sec:intro}

Nowadays, Continual Learning (CL) \cite{DBLP:books/sp/98/Ring98,mccloskey1989catastrophic} is proposed to mimic the human learning process which maintains old knowledge when acquiring new skills and knowledge. 
However, existing CL approaches will forget previous knowledge when they learn new tasks, \ie 
catastrophic forgetting.
Therefore, several branches of algorithms have been proposed to address this problem \cite{DBLP:conf/eccv/AljundiBERT18, DBLP:conf/icml/ZenkePG17, DBLP:journals/pami/LiH18a,DBLP:journals/corr/RusuRDSKKPH16,DBLP:conf/iclr/VeniatDR21,DBLP:conf/cvpr/RebuffiKSL17,DBLP:conf/iclr/SprechmannJRPBU18,DBLP:conf/iclr/ChaudhryRRE19,DBLP:conf/mm/ChenZSG22}. 
However, all these methods just satisfy the supervised learning paradigm where class labels for data points are given. 
And high-quality, class-labeled samples are scarce.

So, Unsupervised Continual Learning (UCL) which has ability to address this issue, receives more and more attention in the community.
Under the UCL paradigm, CURL \cite{DBLP:conf/nips/RaoVRPTH19} is proposed firstly, but, its capability is restricted to tackling complex assignments.
Another work \cite{DBLP:conf/iclr/MadaanYLLH22} broadens the scope of supervised CL approaches in unsupervised paradigm and evaluates the performance of the state-of-the-art unsupervised learning methods, \ie SimSiam \cite{DBLP:conf/cvpr/ChenH21} and BarlowTwins \cite{DBLP:conf/icml/ZbontarJMLD21}. 
However, they only examine these approaches and do not propose a suitable approach for the UCL.

\begin{figure}[t]
\centering
\includegraphics[width=1.0\linewidth]{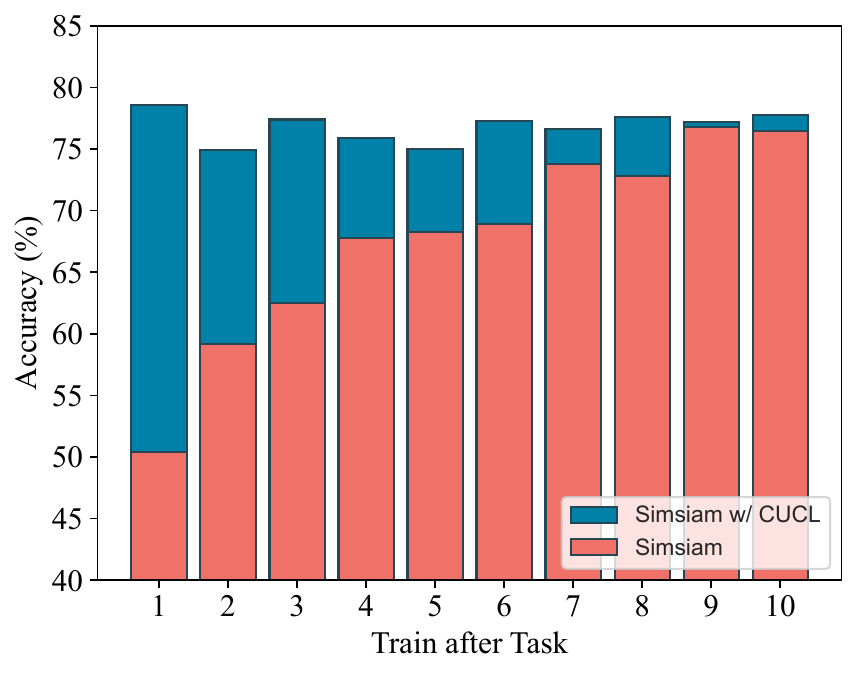}
\caption{The performance of Simsiam \cite{DBLP:conf/cvpr/ChenH21} and Simsiam with CUCL for the first task of Split CIFAR100 \cite{krizhevsky2009learning}. 
Each bar $i$ represents the result of the first task after training on task $i$. 
We observe two main findings. Firstly, the result is suboptimal upon completing the training for the first task. Nonetheless, this result improves as the training progresses. Secondly, we note a considerable improvement in performance when CUCL is integrated with Simsiam.} 
\label{fig:NTG}
\end{figure}

In this paper, we study the UCL paradigm follow \cite{DBLP:conf/iclr/MadaanYLLH22}.
We reveal a phenomenon when training the SimSiam that the results of the first few tasks are suboptimal.
And these results show an upward trend as training continues on the following tasks, as shown in Fig.\ref{fig:NTG}.
However, this occurrence contradicts the practicality of continual learning, which requires the model to address catastrophic forgetting and maintain optimal performance on the current task in hand at the same time.
For example, it is not acceptable to tolerate the poor performance of the AI-Model in classification and anticipate an enhancement in its classification capability after training it on a Vision \& Language task \cite{DBLP:conf/mm/LiGWLXSS19, DBLP:conf/mm/GaoZSLS18, lyu2023adaptive, DBLP:conf/ijcai/ZengZSG22}.
Furthermore, to check whether this phenomenon is a common issue in UCL, we conduct experiments with a series of unsupervised learning methods \cite{DBLP:conf/icml/ChenK0H20,DBLP:conf/nips/GrillSATRBDPGAP20,DBLP:conf/icml/ZbontarJMLD21,DBLP:conf/cvpr/ChenH21,DBLP:conf/nips/CaronMMGBJ20}.
Upon fine-tuning some of these methods, the same phenomenon shows up.
Meanwhile, traditional metrics, namely Average Accuracy (AA) and Backward Transfer (BWT) can no longer provide an accurate assessment of the model's learning ability for each task under UCL paradigm.
Therefore, this paper chooses novel metric which proposed by \cite{DBLP:conf/cvpr/RebuffiKSL17}, and we term it as \textit{Mean Average Accuracy} (MAA), to precisely assess task performance during the training process.

Since the phenomenon can be measured accurately, a solution approach is necessary.
Three assumptions were made after analyzing the results.
1.) The lack of diversity of classes may be the cause.
The small size of classes in each task makes it difficult to obtain a discriminative representation.
Consequently, after conducting experiments by increasing the number of classes in each task, although performances increase, the suboptimal problem remaines, indicating that this is not the primary factor.
2.) The slow convergence of unsupervised methods is deemed to be the cause. 
As a result, methods were trained for an extended period.
This led to a marginal improvement of results, but the suboptimal phenomenon remained, ruling out training time as the key cause.
(refer to Sec. \ref{sec:traingtime} for details)
3.) The lacking diverse information of features in unsupervised methods is believed to be the primary factor.
In addition, we observe that the methods which can only meet the transformed inputs of the same input are more severe about the suboptimal problem.
While they can recognize the inherent features of a class, the inability to interact with other classes impedes discriminative classification.
So we want to confer diverse information (about other classes) for representations to guide the model to learn invariable information about the class contained in each task.
\textit{Vector Quantization} (VQ) is used to achieve this objective.
Clustering features extracted by backbone produces centroids, or codewords, within a codebook. 
These codewords contain diverse information about various classes.
The final representation created by these codewords absorbs sufficient diversity to create a discriminative class boundary.
Experiments were conducted by this way to inject diverse information into representations for unsupervised learning techniques.
The findings suggest that it enhances convergence, resulting in a 15\% improvement (refer to Sec. \ref{Forloss} for details).

% Through these experiments, diverse information about other samples is crucial to solve the suboptimal problem, under the UCL paradigm.
% It is essential to inject diverse information without altering the architecture of the methods.
% Therefore, we select the representation extracted by these methods to facilitate contrastive learning for interacting.

Furthermore, we contend that distinct segments of a representation contain distinct local patterns.
Focusing diversity on these segments rather than the entire representation will achieve more fine-grained diverse.
Meanwhile, the fundamental principle of \textit{Product Quantization} (PQ) is to decompose a feature vectors in high-dimensional space into multiple sub-vectors.
The ultimate representation is a rearrangement of codewords that contains fine-grained information.
This process is highly consistent with our idea.
Consequently, in this paper, we propose a \textbf{C}odebook for \textbf{U}nsupervised \textbf{C}ontinual \textbf{L}earning (CUCL), which employs PQ to quantize representations learned from unsupervised learning methods.
Then, we apply contrastive learning to maximize the cross-similarity between the original representations and the quantized ones of another branch, while minimizing similarity with other samples to learn invariable knowledge for promoting classification boundary.

Although, the proposed CUCL can resolve the suboptimal problem.
The primary issue in CL, catastrophic forgetting, persists. 
Fortunately, the CUCL enables the codewords to learn the features of each task, and these codewords can serve as proxies for selecting representative samples. 
Consequently, an algorithm based on the codewords is proposed to mitigate catastrophic forgetting by selecting the closest samples with codewords for rehearsal.

% We argue that the reason for NTF is lacking of the positive and nagative samples.
% So the model can not learn the discriminative representations.
% Therefore, we use a Codebook to split the representation learned after various contrastive learning methods into segments and apply the soft quantization on them.

% \textit{Product Quantization} (PQ) to quantize the output of each branch of contrastive learning network, then maximizes the cross-similarity between the origin representation and the product quantized representation of another branch. This strategy encourages the network to learn the discriminative image contents representations.

In summary, the proposed CUCL consists of two parts. 
One part utilizes PQ to quantize representations and perform contrastive learning, resolving the suboptimal problem. 
The other part selects the closest samples for rehearsal as a means of mitigating catastrophic forgetting. 
We conducte extensive experiments to evaluate the efficacy of our proposed method on multiple continual learning benchmarks, including \textbf{CIFAR100}, \textbf{TinyImageNet}, and \textbf{MiniImageNet}, by testing various supervised CL methods and their variants on UCL.
The results prove the ability of our method in enhancing these approaches. 
For, instance, we have observed average improvements of 9.08\%, 7.64\% and 6.49\% on the three datasets over SI, DER, LUMP which are impletemeneted on Simsiam, in term of MAA.
Additionally, we evaluate the effectiveness of our method by conducting more experiments on state-of-the-art unsupervised learning methods. 
The integration of our CUCL results in substantial enhancements.

To summarize, our main contributions are three-fold:
\begin{itemize}
\item A study under Unsupervised Continual Learning (UCL) paradigm has been conducted, which uncovers a phenomenon that the efficacy on the first few tasks of certain unsupervised learning methods is limited by the diverse information contained in features. 
% Further, a new metric is proposed to address the limitations of 
% traditional metrics in assessing overall performance, named Mean Average Accuracy (MAA).

\item We have introduced a Codebook for Unsupervised Continual Learning (CUCL), which is a plug-and-play approach towards enabling networks to acquire discriminative information through quantized representation. 
Furthermore, we establish a rehearsal algorithm based on the codebook to mitigate the catastrophic forgetting issue.

\item To demonstrate the effectiveness of our method, an extensive experimental evaluation of several benchmark datasets has been conducted.
The marginal improvement confirms our method's ability to solve the suboptimal problem and alleviate catastrophic forgetting.
\end{itemize}

\section{Related Work}
\label{sec:relatedwork}

\subsection{Continual Learning}
With the growing interest in continual learning, many methods have been proposed to address the catastrophic forgetting problem. 
They can be grouped into three broad categories: regularization approaches \cite{DBLP:conf/eccv/AljundiBERT18, DBLP:journals/corr/KirkpatrickPRVD16, DBLP:conf/icml/ZenkePG17, DBLP:journals/pami/LiH18a}, parameter isolation methods \cite{DBLP:conf/icml/SerraSMK18,DBLP:journals/corr/RusuRDSKKPH16,DBLP:conf/iclr/VeniatDR21}, and memory-based approaches \cite{DBLP:conf/nips/Lopez-PazR17,DBLP:conf/cvpr/RebuffiKSL17,DBLP:conf/iclr/SprechmannJRPBU18,DBLP:conf/iclr/AyubW21,DBLP:conf/iclr/ChaudhryRRE19}.
The \textit{Regularization approaches} focus on curing a continual learning network of its catastrophic forgetting by introducing an extra regularization term in the loss function.
LwF \cite{DBLP:journals/pami/LiH18a} mitigated forgetting by using knowledge distillation and transferring knowledge, and used previous model output as soft labels for the previous task.
Besides, EWC \cite{DBLP:journals/corr/KirkpatrickPRVD16} was the first method to penalize the changes to important parameters during training of later tasks.
SI \cite{DBLP:conf/icml/ZenkePG17} efficiently estimated importance weights during training.
MAS \cite{DBLP:conf/eccv/AljundiBERT18} computed the importance of the parameters of a neural network in an unsupervised and online manner.
The basic idea of \textit{Parameter isolation methods} is to directly add or modify the model structure.
HAT \cite{DBLP:conf/icml/SerraSMK18} applied the mask on previous task parts during new task training, and this process is imposed at the unit level.
PNN \cite{DBLP:journals/corr/RusuRDSKKPH16} added a network to each task and lateral connections to the network of the previous task while freezing previous task parameters.
MNTDP \cite{DBLP:conf/iclr/VeniatDR21} proposed a modular layer network approach, whose modules represent atomic skills that can be composed to perform a certain task and provides a learning algorithm to search the modules to combine with. These strategies may work well, but they are computationally expensive and memory intensive.
For \textit{Memory-based approaches}, catastrophic forgetting is avoided by storing data from previous tasks and training them together with data from the current task.
Some methods \cite{DBLP:conf/nips/Lopez-PazR17,DBLP:conf/cvpr/RebuffiKSL17,DBLP:conf/iclr/SprechmannJRPBU18,DBLP:conf/iclr/ChaudhryRRE19} used replayed samples from previous tasks to constrain the parameters’ update when learning the new task.
For example, iCaRL \cite{DBLP:conf/cvpr/RebuffiKSL17} selected a subset of exemplars that are the best approximate class means in the learned feature space.
During training on a new task of EEC \cite{DBLP:conf/iclr/AyubW21}, reconstructed images from encoded episodes were replayed to avoid catastrophic forgetting.
Although these methods have achieved remarkable performance, they are just designed for supervised continual learning. 
Our approach is designed for the unsupervised setting which is more realistic. 

\subsection{Unsupervised Representational Learning}
Recently there has been steady progress in unsupervised representational learning. 
Many methodologies \cite{DBLP:conf/cvpr/WuXYL18,DBLP:conf/iclr/HjelmFLGBTB19,DBLP:conf/cvpr/YeZYC19,DBLP:conf/eccv/TianKI20,DBLP:conf/cvpr/He0WXG20,DBLP:conf/icml/ChenK0H20,DBLP:conf/icml/ZbontarJMLD21,DBLP:conf/nips/GrillSATRBDPGAP20,DBLP:conf/cvpr/ChenH21, DBLP:conf/nips/CaronMMGBJ20,DBLP:conf/cvpr/HeCXLDG22} have been proposed for un-/self-supervised learning.
SimCLR \cite{DBLP:conf/icml/ChenK0H20} used the other samples coexisting in the current batch as negative samples, so it worked well when equipped with a large batch size.
MoCo \cite{DBLP:conf/cvpr/He0WXG20} applied a queue of negative samples as a dynamic look-up dictionary and a momentum encoder which is proposed to maintain consistency.
BYOL \cite{DBLP:conf/nips/GrillSATRBDPGAP20} was a Siamese \cite{DBLP:conf/nips/BromleyGLSS93} in which one branch is a momentum encoder. 
It directly predicted the network representation of one view from another view.
BarlowTwins \cite{DBLP:conf/icml/ZbontarJMLD21} avoided collapse by measuring the cross-correlation matrix between the outputs of two identical networks, and making it as close to the identity matrix as possible.
SwAV \cite{DBLP:conf/nips/CaronMMGBJ20} incorporateed online clustering into Siamese networks.
In another recent line of work, the network architecture and parameter updates of SimSiam \cite{DBLP:conf/cvpr/ChenH21} are modified to be asymmetric such that the parameters are only updated using one branch.

\begin{figure*}[t]
\centering
\includegraphics[width=1.0\linewidth]{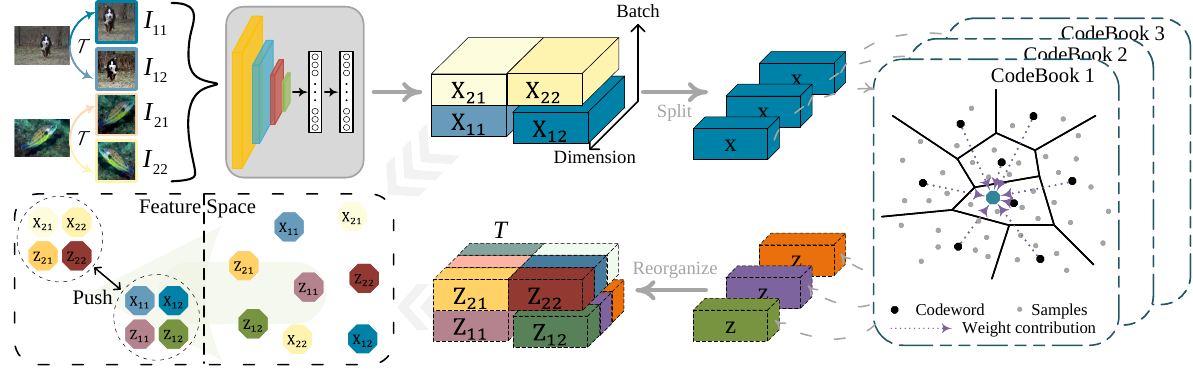}
\caption{An overview of the proposed CUCL.
    The samples are augmented using data augmentations $\mathcal{T}$ and then passed on to the Backbone to obtain the original feature representation $\mathbf{X} \in \mathcal{R}^D$.
    The representations are subjected to the traditional unsupervised learning loss $\mathcal{L}_{unsup}$. 
    Subsequently, the original representations are divided into subvectors $\mathbf{x} \in \mathcal{R}^{D/M}$.
    The soft quantizer is applied to the subvectors $\mathbf{x}$ to obtain quantized subvectors $\mathbf{z}$.
    Then, the quantized subvectors $\mathbf{z}$ are reorganized into a representation $\mathbf{Z}$, which leads to enhanced diversity and robustness.
    The cross contrastive loss $\mathcal{L}_{cucl}$ is imposed on both the original representations $\mathbf{X}$ and the quantized $\mathbf{Z}$ in the subsequent stage.
    Finally, the model is optimized by the final loss: $\mathcal{L} = \mathcal{L}_{unsup} + \mathcal{L}_{cucl}$.}
\label{fig:architecture}
\end{figure*}

\section{Methodology}
In this section, we describe the settings of UCL and our methodology.
Specifically, we present the problem formulation and the metrics used in traditional continual learning in Section \ref{preliminary}.
Section \ref{MAA} explains the new metric to measure the performance along the training process.
Then, Section \ref{CUCL} and Section \ref{rehearsal}  provide the details of the proposed method and explain how to select rehearsal samples.

\subsection{Preliminary and Metrics}
\label{preliminary}
In the setup of unsupervised continual learning, a series of $T$ tasks are learned sequentially. 
We denote a task by its task descriptor, $\tau \in \{1,2,...,T\}$ and its corresponding dataset $\textit{D}_\tau = \{{(I_{\tau,i})}_{i=1}^{N_\tau}\}$ which has $N_\tau$ samples from an i.i.d. distribution.
The $I_{\tau,i}$ is the input without supervised label. 
Furthermore, the task boundaries are available during both training and testing stage, \ie the task-incremental setting.
The UCL seeks to train a model $F = \{\theta_{F}\}$ to learn knowledge on a sequence of tasks without forgetting the knowledge learned on previous tasks, where $\theta_{F}$ represents the weights of the neural network.

The widely used evaluation metrics in continual learning are: Average Accuracy (ACC), and Backward Transfer (BWT). 
Formally, ACC and BWT are defined as:
\begin{align}
ACC &=\frac{1}{T} \sum_{i=1}^TA_{T,i},\\
BWT &=\frac{1}{T-1} \sum_{i=1}^{T-1}A_{T,i}-A_{i,i},
\end{align}
where $A_{T, i}$ is the performance on $\tau=i$ task after training on the $\tau=T$ task.

In addition, we use Mean Average Accuracy (MAA) \cite{DBLP:conf/cvpr/RebuffiKSL17}.
Formally, MAA is defined as:
\begin{equation}
MAA=\frac{1}{T} \sum_{j=1}^T (\frac{1}{j} \sum_{i=1}^j A_{j,i}).
\end{equation}
The MAA is calculated as the mean of the average accuracy for tasks at each training point that the model has encountered.
A high MAA corresponds to a continual learning model that consistently maintains a high accuracy throughout the training process.

\subsection{Codebook for Unsupervised Continual Learning}
\label{CUCL}
In this section, we propose the Codebook for Unsupervised Continual Learning (CUCL) to address the suboptimal phenomenon prevalent in unsupervised continual learning. 
Fig. \ref{fig:architecture} presents an overview of the proposed CUCL pipeline. 
Prior to detailing the technical aspects, we offer a high-level understanding of the workings of CUCL. 
Our approach aims to facilitate model learning of discriminative knowledge by fostering diversity into representations. 
Furthermore, we contend that different segments of representation harbor distinctive local patterns. 
To this end, we first quantize the features into different segments based on their local patterns. 
Features assigned to the same codeword contain the same local information, whereas features with different information have different codewords. 
By reconstructing the quantized segments, we accumulate more information and enable more fine-grained analysis. 
In summary, we transform the original feature into a combination of different local pattern features to enhance model learning of more discriminative feature.

First of all, we briefly introduce the unsupervised learning methods working on Siamese networks.
These methods operate on a pair of embeddings extracted from transformed images.
More specifically, they produce two distorted views via data augmentations $\mathcal{T}$ for all images of a batch sampled from $\textit{D}_\tau$.
The distorted views are then fed to the model $F$, producing two batches of embeddings $\mathbf{X}_{a}$ and $\mathbf{X}_{b}$ respectively, each $\mathbf{X} \in \mathcal{R}^D$.
The different unsupervised learning loss will be applied to $\mathbf{X}_{a}$ and $\mathbf{X}_{b}$ to produce the optimization objective, $\mathcal{L}_{unsup}$.

After we get the features $\mathbf{X}$ extracted by the $F$, following the method in \cite{DBLP:journals/ijcv/YuMFJY20,DBLP:conf/iccv/JangC21}, we quantize them using the soft quantization model $Q(\mathbf{X};\theta_{Q})$.
This quantization method can solve the infeasible derivative calculation of hard assignment quantization and be trained in an end-to-end manner.
The quantization model $Q$ contains $M$ codebooks $\left \{ \mathcal{B}_1,...\mathcal{B}_M \right \}$, where each codebook $\mathcal{B}_i$ has $K$ codeword $\mathbf{c} \in \mathcal{R}^{D/M}$ as $\mathcal{B}_i = \left \{ \mathbf{c}_{i1},...,\mathbf{c}_{iK}  \right \} $. 
We split the features $\mathbf{X}$ into $M$ subvectors $[\mathbf{x}_1,...,\mathbf{x}_M]$ in the feature space where $\mathbf{x}_i \in \mathcal{R}^{D/M}$ is a subvector.
And then, we use the soft quantization model to quantize these subvectors.
First of all, we compute the distance between the subvector $\mathbf{x}_i$ and the codeword:
\begin{equation}
\label{eq.dis}
dis(\mathbf{x}_{i},\mathbf{c}_{ik}) = \left \| \mathbf{x}_{i}-\mathbf{c}_{ik} \right \|^2_2,
\end{equation}
where $\left \| \cdot \right \| ^2_2$ represents the squared Euclidean distance.
The quantization process for each feature subvector is defined below:
\begin{equation}
\mathbf{z}_{i} = \sum_{k}^{K} \frac{exp(-dis(\mathbf{x}_{i},\mathbf{c}_{ik}) /\tau_q)}{\sum_{k'}^{K} exp(-dis(\mathbf{x}_{i},\mathbf{c}_{ik'})/\tau_q  )}\mathbf{c}_{ik} ,
\end{equation}
where $\tau_q$ is a temperature parameter to scale the input.
The contribution of each codeword is associated with the distance from the codeword to the subvector, where the closest codeword has the biggest contribution and vice versa.
By applying this quantization to $\mathbf{X}_{a}$ and $\mathbf{X}_{b}$, we can get the quantized features $\mathbf{Z}_{a}$ and $\mathbf{Z}_{b}$ by combining the quantized subvectors $\mathbf{z}$.
According to this soft quantization, the information contained in each codeword is integrated into the final quantization representation, producing a more diverse representation to learn discriminative feature.

\begin{table*}
\caption{The quantitative results of supervised baselines and their unsupervised variants on Resnet-18 architecture with KNN classifier \cite{DBLP:conf/cvpr/WuXYL18}. While \textit{w}/ CUCL is the adaptation of our method to other methodologies. There are considerable improvements when our CUCL plugged into each baseline.}
\label{tab:SupervisedResult}
\renewcommand\arraystretch{1.3}
\renewcommand\tabcolsep{4.0pt}
\centering
\resizebox{\linewidth}{!}{
    \begin{tabular}{clccccccccc}
        \hline
        \multirow{2}{*}{Setting} & \multirow{2}{*}{Method} & \multicolumn{3}{c}{Split CIFAR100} & \multicolumn{3}{c}{Split TinyImageNet} & \multicolumn{3}{c}{Split MiniImageNet} \\ \cline{3-11}
        &  & MAA & AA & BWT & MAA & AA & BWT & MAA & AA & BWT \\ \hline \hline
        \multirow{3}{*}{Supervised} & SI \cite{DBLP:conf/icml/ZenkePG17} & 59.17 & 55.77 & -32.00 & 52.12 & 44.42 & -35.27 & 50.12 & 44.97 & -34.52 \\
        & A-GEM \cite{DBLP:conf/iclr/ChaudhryRRE19} & 58.10 & 54.77 & -31.24 & 52.00 & 48.74 & -29.67 & 50.84 & 48.09 & -30.32 \\
        & DER \cite{DBLP:conf/nips/BuzzegaBPAC20} & 72.48 & 67.97 & -20.09 & 60.83 & 56.74 & -21.40 & 58.52 & 54.49 & -23.40 \\ \hline
        
        \multirow{6}{*}{Simsiam} & SI & 69.75 & 74.32 & -2.66 & 51.37 & 58.80 & -0.98 & 54.79 & 63.24 & 0.30 \\ 
        & \CC \textit{w}/ CUCL 
        & \CC 75.79$_{\textcolor{red}{+6.04}}$ 
        & \CC 76.68$_{{+2.36}}$ 
        & \CC -5.12$_{{-2.46}}$
        & \CC \textbf{62.45}$_{\textcolor{red}{+11.08}}$ 
        & \CC 64.48$_{{+5.68}}$ 
        & \CC -4.49$_{{-3.51}}$ 
        & \CC 64.91$_{\textcolor{red}{+10.12}}$ 
        & \CC 67.05$_{{+3,81}}$ 
        & \CC -4.44$_{{-4.74}}$ \\ \cdashline{2-11}
        
        & DER & 73.05 & 75.31 & -4.04 & 51.01 & 58.66 & -0.82 & 56.78 & 63.72 & -0.61 \\
        & \CC \textit{w}/ CUCL 
        & \CC \textbf{76.32}$_{\textcolor{red}{+3.27}}$
        & \CC \textbf{77.24}$_{{+1.93}}$
        & \CC -4.39$_{{-0.35}}$
        & \CC 62.41$_{\textcolor{red}{+11.4}}$ 
        & \CC \textbf{65.10}$_{{+6.44}}$ 
        & \CC -4.04$_{{-3.22}}$ 
        & \CC \textbf{65.03}$_{\textcolor{red}{+8.25}}$ 
        & \CC \textbf{67.15}$_{{+3.43}}$ 
        & \CC -4.42$_{{-3.81}}$   \\ \cdashline{2-11}
        
        & LUMP \cite{DBLP:conf/iclr/MadaanYLLH22} & 63.95 & 69.83 & \textbf{3.60} & 53.60 & 61.12 & \textbf{5.93} & 52.36 & 59.54 & \textbf{4.52} \\
        & \CC \textit{w}/ CUCL 
        & \CC 72.19$_{\textcolor{red}{+8.24}}$ 
        & \CC 74.78$_{{+4.95}}$
        & \CC -0.82$_{{-4.42}}$
        & \CC 58.51$_{\textcolor{red}{+4.91}}$ 
        & \CC 59.42$_{{-1.7}}$ 
        & \CC -2.80$_{{-8.73}}$ 
        & \CC 58.67$_{\textcolor{red}{+6.31}}$ 
        & \CC 60.25$_{{+0.71}}$
        & \CC -2.36$_{{-6.98}}$ \\ \hline
        
        \multirow{6}{*}{Barlow twins} & SI & 73.13 & 73.52 & -8.16 & 60.39 & 59.72 & -8.73 & 61.53 & 61.97 & -7.91 \\
        & \CC \textit{w}/ CUCL 
        & \CC 74.40$_{\textcolor{red}{+1.27}}$ 
        & \CC 74.93$_{{+1.41}}$
        & \CC -7.68$_{{+0.48}}$
        & \CC 61.59$_{\textcolor{red}{+1.2}}$ 
        & \CC 62.44$_{{+2.72}}$
        & \CC -6.11$_{{+2.62}}$
        & \CC 62.67$_{\textcolor{red}{+1.14}}$ 
        & \CC 64.50$_{{+2.53}}$ 
        & \CC -6.61$_{{+1.3}}$ \\ \cdashline{2-11}
        
        & DER & 73.15 & 73.46 & -7.63 & 60.40 & 60.04 & -8.82 & 61.79 & 62.47 & -7.93 \\
        & \CC \textit{w}/ CUCL 
        & \CC 74.62$_{\textcolor{red}{+1.47}}$ 
        & \CC 75.47$_{{+2.01}}$
        & \CC -7.16$_{{+0.47}}$
        & \CC 61.57$_{\textcolor{red}{+1.17}}$ 
        & \CC 60.88$_{{+0.84}}$
        & \CC  -8.82$_{{+0.00}}$
        & \CC 63.10$_{\textcolor{red}{+1.31}}$ 
        & \CC 63.82$_{{+1.35}}$
        & \CC -7.64$_{{+0.29}}$ \\ \cdashline{2-11}
        
        & LUMP & 67.18 & 66.73 & -6.33 & 55.26 & 57.78 & -2.62 & 54.91 & 57.66 & -2.63 \\
        & \CC \textit{w}/ CUCL 
        & \CC 72.16$_{\textcolor{red}{+4.98}}$ 
        & \CC 73.66$_{{+6.93}}$
        & \CC -3.88$_{{+2.45}}$
        & \CC 59.43$_{\textcolor{red}{+4.17}}$  
        & \CC 61.58$_{{+3.8}}$ 
        & \CC -3.04$_{{-0.42}}$  
        & \CC 58.60$_{\textcolor{red}{+3.69}}$ 
        & \CC 61.80$_{{+4.14}}$
        & \CC -1.97$_{{+0.66}}$ \\ \hline
    \end{tabular}
}
\end{table*}

Then we apply a cross contrastive learning inspired by traditional contrastive learning to compare the $\mathbf{X}$ and $\mathbf{Z}$ of different views.
Same as the contrastive learning, we treat the $\mathbf{X}$ and $\mathbf{Z}$ as positive if they are generated from the same image, whereas negative if originated from the different ones.
So we can apply contrastive learning loss between the $\mathbf{X}$ and $\mathbf{Z}$ according to the positive pair of examples $(i,j)$:
\begin{equation}
\mathcal{L}_{cucl} = -log\frac{exp(Cos(\mathbf{X}_i,\mathbf{Z}_j)/\tau_l)}{\sum_{n=1}^{N_B} \mathbb{1}_{[n \neq j]} exp(Cos(\mathbf{X}_i,\mathbf{Z}_{n})/\tau_l)} ,
\end{equation}
where $Cos$ represents cosine function, $\tau_l$ is a non-negative temperature parameter, and the $\mathbb{1}_{[n'\neq j]} \in {0,1}$ denotes the indicator which equates to 1 iff $n \neq j$. 
% When the $j$ is odd, the $n'$ is $2n-1$, otherwise $2n$.
% This avoids the redundancy between $\mathbf{X}$ and $\mathbf{Z}$ which are similar to each other.

The overall loss combines the traditional unsupervised loss and our CUCL loss, as
\begin{equation}
\mathcal{L} = \mathcal{L}_{unsup} + \mathcal{L}_{cucl}.
\end{equation}
% The key steps of the algorithm are listed in Algorithm \ref{algorithm1}.

% \begin{algorithm}[t]
% \caption{Algorithm for CUCL.}
% \label{algorithm1}
% \begin{algorithmic}[1]
    %     \Require
    %           Backbone $F$; Quantization model $Q$; Task sequence $T$; Learning rate $\alpha$; Train epoch $E$; Transform $T$; Unsupervised learning loss $Con$. 
    %     \Ensure
    %       Backbone $F$; Quantization model $Q$.
    %     \State Initialize Model $\Phi$: $\Phi \leftarrow \Phi_0$.
    %     \For {$\tau=0,\ldots,\left | T \right | $}
    %     	\For {$e_\tau=0,\ldots,E$} 
    %         	\State $B \sim D_{\tau}$  
    %         	\State $B_{a},B_{b} \sim T(B)$ 
    %         	\State	$X_a,X_b \leftarrow F(B_{a}),F(B_{b})$
    %      	    \State $\mathcal{L}_{unsup} \leftarrow  Con(X_a,X_b)$
    %      	    \State $Z_a,Z_b \leftarrow Q(X_{a}),Q(X_{b})$
    %      	    \State $\mathcal{L}_{cucl} \leftarrow  Con(X_a,X_b,Z_a,Z_b)$
    %      	    \State $\theta_F \leftarrow \theta_F - \alpha  ( \frac{\partial \mathcal{L}_{unsup}}{\partial \theta_F} + \frac{\partial \mathcal{L}_{cucl}}{\partial \theta_F}) $
    %      	    \State $\theta_Q \leftarrow \theta_Q - \alpha   \frac{\partial \mathcal{L}_{cucl}}{\partial \theta_F} $
    %     	\EndFor
    %     \EndFor
    %     \State \Return $F,Q$
    % \end{algorithmic}
% \end{algorithm}

\subsection{Codebook Rehearsal}
\label{rehearsal}
With the cross quantized contrastive learning, the model works well on the tasks at hand.
The problem that the first few tasks are suboptimal is ameliorated.
However, the catastrophic forgetting remains.
Since the codewords represent the clustered centroid of the subvectors, they can be seen as the proxies of the subvectors and the carrier of the information.
Specifically, the codeword $\mathbf{c}_{ic}$ which is close to subvector $\mathbf{x}_i$ serves as the proxy of this subvector.
We use these codewords to choose some representative samples for rehearsal.
Simply put, since the weight contribution is related to the distance in Eq. \ref{eq.dis}, we use this distance as a clue to choose samples.
We estimate the distance between each subvector $\mathbf{x}_i$ of $\mathbf{X}$ and its proxy $\mathbf{c}_{ic}$ and sum them up to the final distance:
\begin{equation}
dis(X) = \sum_{i}^{M} \left \| \textbf{x}_i - \textbf{c}_{ic}   \right \|_2^2.
\end{equation}
In this paper, the first $S$ samples which have the furthest distance will be selected.
Because we regard these samples as the most difficult data points.
We can not only recall the knowledge about the task learned before through these buffers, but also retrain the samples to learn more information about the task.

\section{Experiment}
In this section, to evaluate the effectiveness of our proposed method CUCL, we validate it on a variety of continual learning benchmarks. 
Additionally, we perform ablation studies to explore the usefulness of the different components.
\subsection{Experimental setting}
\paragraph{Datasets}
We conduct experiments on various continual learning benchmarks, including \textbf{Split CIFAR100} \cite{krizhevsky2009learning}, \textbf{Split TinyImageNet}, \textbf{Split MiniImageNet}. 
The Split CIFAR100 is constructed by randomly splitting 100 classes of CIFAR100 into 10 tasks, where each class includes 500 training samples and 100 testing samples.
The Split TinyImageNet is a variant of the ImageNet dataset \cite{DBLP:conf/cvpr/DengDSLL009} comprising 10 randomized classes out of the 100 classes for each task, where the images are sized 64 x 64 pixels.
Finally, Split MiniImageNet \cite{DBLP:conf/nips/VinyalsBLKW16} is a dataset created by dividing the 100 classes of the ImageNet into 10 sequential tasks, each consisting of 10 classes.
Each class includes 500 training samples and 100 testing samples.
Each image in the mentioned datasets has a size of 84 × 84 pixels.

\begin{table*}
\caption{The quantitative result of each unsupervised methods on Resnet-18 architecture with KNN classifier. 
    While \textit{w}/ CUCL is the adaptation of our method to other unsupervised learning methodologies. 
    The best results are highlighted in bold.}
\label{tab:mainResult}
\renewcommand\arraystretch{1.3}
\renewcommand\tabcolsep{4.0pt}
\centering
\resizebox{\linewidth}{!}{
    \begin{tabular}{lccccccccc}
        \hline
        \multirow{2}{*}{Method}  & \multicolumn{3}{c}{Split CIFAR100} & \multicolumn{3}{c}{Split TinyImageNet} & \multicolumn{3}{c}{Split MiniImageNet} \\ \cline{2-10} 
        & MAA & AA & BWT & MAA & AA & BWT & MAA & AA & BWT \\ \hline \hline
        
        Simsiam \cite{DBLP:conf/cvpr/ChenH21} & 67.48 & 73.87 & \textbf{-1.20} & 50.47 & 57.18 & -2.62 & 57.33 & 64.36 & -0.92 \\ 
        \rowcolor{mygray}\ \textit{w}/ CUCL & 76.07$_{\textcolor{red}{+8.59}}$ & 75.81$_{+1.94}$ & -6.26$_{-5.06}$ & 63.23$_{\textcolor{red}{+12.76}}$ & 64.42$_{+7.24}$ & -5.18$_{-2.56}$ & \textbf{64.61$_{\textcolor{red}{+7.28}}$} & 66.17$_{+1.81}$ & -5.30$_{-4.38}$ \\ \cdashline{0-9}
        
        BYOL \cite{DBLP:conf/nips/GrillSATRBDPGAP20} & 75.45 & \textbf{77.80} & -1.40 & 56.90 & 63.98 & \textbf{1.16} & 57.75 & 66.56 & \textbf{3.96} \\ 
        \rowcolor{mygray}\ \textit{w}/ CUCL & \textbf{76.78$_{\textcolor{red}{+1.33}}$} & 77.67$_{-0.13}$ 
        & -4.19$_{-2.79}$ 
        & \textbf{63.90$_{\textcolor{red}{+7}}$} 
        & \textbf{65.66$_{+1.68}$} 
        & -4.58$_{-5.74}$
        & 64.50$_{\textcolor{red}{+7.28}}$ 
        & \textbf{66.99$_{+0.43}$} 
        & -3.74$_{-7.7}$ \\ \cdashline{0-9}
        
        Barlow-twins \cite{DBLP:conf/icml/ZbontarJMLD21} & 72.91 & 72.96 & -8.47 & 59.90 & 59.76 & -8.22 & 61.26 & 63.32 & -6.92 \\ 
        \rowcolor{mygray}\ \textit{w}/ CUCL & 75.29$_{\textcolor{red}{+2.38}}$ & 75.18$_{+2.22}$ & -8.82$_{-0.35}$ & 63.12$_{\textcolor{red}{+3.22}}$ & 63.58$_{+3.82}$ & -7.27$_{+0.95}$ & 64.31$_{\textcolor{red}{+3.05}}$ & 66.16$_{+2.84}$ & -6.56$_{+0.36}$ \\ \cdashline{0-9}
        
        SimCLR \cite{DBLP:conf/icml/ChenK0H20} & 74.39 & 74.14 & -6.58 & 63.49 & 63.24 & -6.56 & 63.15 & 65.30 & -4.71 \\ 
        \rowcolor{mygray}\ \textit{w}/ CUCL & 75.38$_{\textcolor{red}{+0.99}}$ & 75.88$_{+1.74}$ & -5.23$_{+1.35}$ & 63.81$_{\textcolor{red}{+0.42}}$ & 64.14$_{+0.9}$ & -5.36$_{+1.2}$ & 63.72$_{\textcolor{red}{+0.57}}$ & 64.77$_{-0.53}$ & -5.09$_{-0.38}$ \\ \cdashline{0-9}
        
        SwAV \cite{DBLP:conf/nips/CaronMMGBJ20} & 73.18 & 74.44 & -4.20 & 58.52 & 62.94 & -0.76 & 62.03 & 62.66 & -4.67 \\
        \rowcolor{mygray}\ \textit{w}/ CUCL & 74.77$_{\textcolor{red}{+1.59}}$ & 74.86$_{+0.42}$ & -6.39$_{-2.19}$ & 61.95$_{\textcolor{red}{+3.43}}$ & 62.84$_{-0.1}$ & -5.40$_{-4.64}$ & 63.22$_{\textcolor{red}{+1.19}}$ & 64.29$_{+1.63}$ & -5.63$_{-0.96}$ \\ \hline
        % \cdashline{0-9}
        % MoCoV2 \cite{DBLP:conf/cvpr/He0WXG20} &  &  &  &  &  &  &  &  &  \\ 
        % \rowcolor{mygray}\ \textit{w}/ CUCL &  &  &  &  &  &  &  &  &  \\ \hline
    \end{tabular}
}
\end{table*}

\paragraph{Baselines and Training setup}
Firstly, we compare our method with state-of-the-art continual learning methods, includes \textbf{SI} \cite{DBLP:conf/icml/ZenkePG17}, \textbf{A-GEM} \cite{DBLP:conf/iclr/ChaudhryRRE19}, \textbf{DER} \cite{DBLP:conf/nips/BuzzegaBPAC20}, \textbf{LUMP} \cite{DBLP:conf/iclr/MadaanYLLH22} and their unsupervised variants.
Secondly, we compare our method with state-of-the-art unsupervised learning methods, including \textbf{Simsiam} \cite{DBLP:conf/cvpr/ChenH21}, \textbf{Barlow twins} \cite{DBLP:conf/icml/ZbontarJMLD21}, \textbf{BYOL} \cite{DBLP:conf/nips/GrillSATRBDPGAP20}, \textbf{SimCLR} \cite{DBLP:conf/icml/ChenK0H20} and \textbf{SwAV} \cite{DBLP:conf/nips/CaronMMGBJ20}.
All these methods employ a common underlying architecture, a Siamese \cite{DBLP:conf/nips/BromleyGLSS93} network structure.
Our approach can be integrated into these networks since we also utilize the same network structure.

We implement CUCL and reproduce the results of all these methods from the released code of \cite{DBLP:conf/iclr/MadaanYLLH22}.
We use the same hyperparameters as \cite{DBLP:conf/iccv/ChenXH21,DBLP:conf/iclr/MadaanYLLH22}.
The K-Nearest Neighbors (KNN) \cite{DBLP:conf/cvpr/WuXYL18} classifier is used to evaluate the representations obtained from the unsupervised learning. 
Resnet18 \cite{DBLP:conf/cvpr/HeZRS16} is employed as the backbone and trained for 200 epochs in each task with a batch size of 256 in the UCL pardigm.
In the supervised paradigm, we follow the setting introduced by \cite{DBLP:conf/iclr/MadaanYLLH22} and train the methods for 50 epochs with a batch size of 32.
We set the temperature parameters for quantization and loss to 5 and 0.5, respectively.
Concerning the quantizer setting in CUCL, we employe 24-bit quantization method, which incorporates 8 codebooks of 8 codewords each with a dimension of 16.
We maintain a uniform learning rate of 0.03 in all the experiments.
Concerning the CUCL rehearsal algorithm, we opt to retain 20 samples per task.
In the main experiments, if there are no special instructions, the CUCL is the combination of CUCL and the Codebook Rehearsal.

\paragraph{Evaluation metrics}
We evaluate the performance on the following metrics: Average Accuracy (ACC), Backward Transfer (BWT) and  Mean Average Accuracy (MAA).

\subsection{Experimental Results}
\subsubsection{Comparison with State of the Arts}
To evaluate the ability of our method to solve the suboptimal problem and mitigate catastrophic forgetting, we compare our method with state-of-the-art supervised methods and its variants under UCL paradigm.
Specifically, we verify our CUCL by incorporating it into these variants.
Following recent work \cite{DBLP:conf/iclr/MadaanYLLH22},
Simsiam and Barlow twins are chosen as unsupervised backbones.
Quantitative results on three datasets are shown in Tab. \ref{tab:SupervisedResult}.

From Tab. \ref{tab:SupervisedResult}, we have following observations:
\textbf{\underline{Firstly}}, the unsupervised variants achieve significantly better performance than supervised methods, as illustrated in \cite{DBLP:conf/iclr/MadaanYLLH22}.
\textbf{\underline{Secondly}},
these variants which are based on Simsiam perform well on three datasets in terms of BWT and AA.
Specifically, about BWT, LUMP achieves positive results as 3.60\%, 5.93\% and 4.52\%.
However, after reviewing the results of each individual task, we observe that the improvements of the first few tasks compensate for the forgetting effect, \ie suboptimal problem.
Consequently, the AA and BWT metrics cannot accurately evaluate the learning ability of the model.
\textbf{\underline{Thirdly}},
in term of MAA, after integrated with CUCL, we have observed average improvements of 9.08\%, 7.64\% and 6.49\% on the three datasets over SI, DER, LUMP, respectively.
These substantial improvements demonstrate the efficiency of our approach in addressing the suboptimal problem.
\textbf{\underline{Fourthly}},
the Barlow twins approach does not suffer the suboptimal problem and performs better than Simsiam. 
Nevertheless, when our CUCL is incorporated with it, there are still 1.2\%, 1.32\% and 4.28\% improvements over SI, DER, LUMP, respectively.
This occurrence indicates that our method is not only capable of addressing the suboptimal problem but also able to learn superior representations that enhance the class boundary.
Moreover, in terms of BWT, the improvements verify the significant efficacy of our approach for mitigating catastrophic forgetting.

\subsubsection{Comparison with Unsupervised Learning Methods}
To verify our method is plug-and-play, we incorporate our CUCL into different unsupervised learning methods.
Quantitative results on three datasets are shown in Tab. \ref{tab:mainResult}.
From Tab. \ref{tab:mainResult}, it's worth noting that, in terms of BWT, the Simsiam and BYOL exhibit remarkable results.
The reason lies in the fact that they also suffer the suboptimal problem (the detail of the learning curve will show in Sec.\ref{MAATracking}).
This issue is addressed by applying our CUCL model, which leads to a significant improvement for the Simsiam and BYOL methods (average 9.54\% and 5.2\% in terms of MAA, respectively).
Specifically, on the CIFAR100 dataset, we observe considerate improvements on Simsiam in terms of MAA (67.48\% \textit{vs.} 76.07\%), showing notable generalizability for CUCL.
In addition, other unsupervised methods employ various techniques to handle negative samples within a batch to solve the suboptimal problem.
Nevertheless, when our approach is  incorporated, significant improvements can still be achieved (\eg 3.43\% of SwAV on TinyImageNet), manifesting the great ability of our CUCL to mitigate the catastrophic forgetting.

Furthermore, as discovered from Tab. \ref{tab:mainResult}, with the task difficulty increasing when training on TinyImageNet and MiniImageNet, results of each methodology suffer degradation.
It is surprising that, in terms of BWT, BYOL even remains positive, indicating that the suboptimal problem is more severe than on CIFAR100.
Additionally, some unsupervised methods (\eg SwAV), which initially perform well on CIFAR100, also suffer from the suboptimal problem.
We speculate it is because discriminative features and segmented quantization are more crucial than simple tasks.
Thus, when our approach is incorporated into these methods, the improvements are more substantial than those achieved in CIFAR100.

% Finally, looking at Table 2 from a general point of view,
% some methods do not suffer from the suboptimal issue (e.g. Barlows-twins, SimCLR), some suffer only slightly (e.g. SwAV), and Simsiam and BYOL suffer the most from it. 
% We think the interaction degree with negative samples may be a potential reason for the appearance of the sub-optimal problem. Specifically, both SimCLR and Barlows-twins are able to interact with negative samples through a batch and a cross-correlation matrix, respectively. 
% Therefore, they do not suffer from sub-optimal issues. In contrast, SwAV indirectly interacts with negative samples via its prototypes, so it suffers only slightly. 
% However, both Simsiam and BYOL can only interact with different views of the same sample (i.e., positive samples), but negative samples. Therefore, they suffer the most. 
% By the above analysis, we conclude that the interaction degree with negative samples is positively correlated with the suboptimal issue.
% In addition, the diverse information of representation is correlated with interaction degree, so our method which conferring diverse information can 

\begin{figure}[t]
\centering
\includegraphics[width=1.0\linewidth]{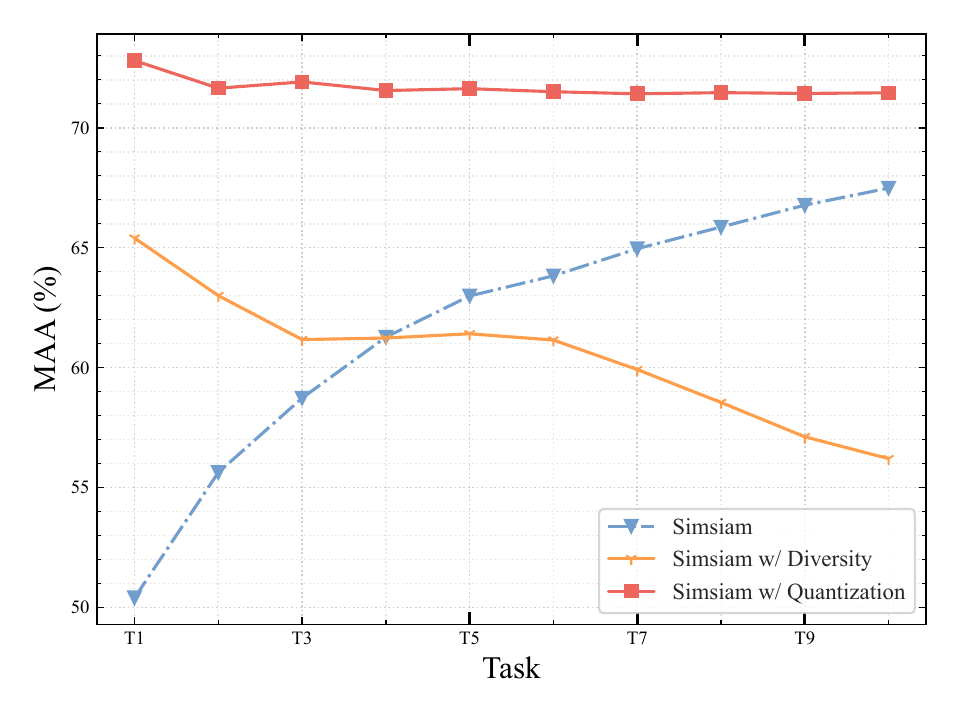}
\vspace{-3em}
\caption{
    The learning curves of MAA.
    These experiments are conducted on the Split CIFAR100 based on the Simsiam methodology. 
    These results demonstrate the necessity of the diversity.}
\vspace{-2em}
\label{fig:Forloss}
\end{figure}

\subsubsection{Ablation Studies}
\paragraph{Analysis of Diversity for Representation}
\label{Forloss}
To confirm the significance of diversity, we conduct experiments on CIFAR100 dataset by chooing Simsiam as unsupervised backbone.
We keep other operations unchanged and only apply quantization to its final representations, which are used for unsupervised loss.
Further, as a comparison base, we record the results of fine-tune.

Initially, we employ a quantizer that only utilize a single codebook to capture the entirety of representations.
This quantizer solely clusters representation to create final representation and does not utilize local patterns in different segments.
The results are shown in Fig. \ref{fig:Forloss}, named Simsiam w/ Diversity.
The figure clearly depicts that the inclusion of diversity substantially mitigate the suboptimal issue, as indicated by a significant improvement in performance from 50\% to 65.4\% on the first task. 
However, along the training process, the inadequacy of using entire representation becomes apparent. 
We speculate it is because the introduction of course diversity impacts the learning process on subsequent tasks as the number of observed classes increases.

This problem can be resolved by utilizing segmented quantized representation, which is validated by Fig. \ref{fig:Forloss}, showing an increase in accuracy of 22.4\% for the first task (\ie Simsiam w/ Quantization). 
Furthermore, upon complete training, the result culminates to 71.46\%. 
These outcomes substantiate our proposal that instilling diversity in distinct representation segments leads to the accumulation of more information.

\begin{table}
\caption{Ablation study about the different codebook settings on the Simsiam w/ CUCL without rehearsal. ‘Cb’ and ‘Cw’ represent the number of codebook and codeword, respectively.}
\label{tab:codebook_results}
\renewcommand\arraystretch{1.3}
\renewcommand\tabcolsep{4.0pt}
\centering
\resizebox{\linewidth}{!}{
    \begin{tabular}{cccccccc}
        \hline
        \multirow{2}{*}{Cb} & \multirow{2}{*}{Cw} & \multicolumn{3}{c}{Split CIFAR100} & \multicolumn{3}{c}{Split TinyImageNet} \\ \cline{3-8}  
        &  & MAA & AA & BWT & MAA & AA & BWT \\ \hline \hline
        1 & 65536 & 71.64 & 71.27 & -10.81 & 61.92 & 64.34 & -4.80 \\ \cdashline{2-8} 
        2 & 256 & 73.72 & 72.95 & -10.31 & 62.82 & 64.34 & -5.40 \\ \cdashline{2-8} 
        4 & 16 & 73.80 & 73.15 & -10.20 & \textbf{62.84} & 63.38 & -6.87 \\ \cdashline{2-8} 
        8 & 4 & \textbf{75.60} & 76.23 & -6.14 & 61.33 & 62.22 & -7.22 \\ \hline
\end{tabular}}
\end{table}

\begin{table}
\caption{This ablation study is to validate the effectiveness of our rehearsal algorithm.
    The experiments are conducted  about different memory size, 0, 20, and 40.
    These results show the effectiveness of our rehearsal algorithm to boost the performances.}
\label{tab:ablation_rehearsal}
\renewcommand\arraystretch{1.3}
\renewcommand\tabcolsep{4.0pt}
\centering
\resizebox{\linewidth}{!}{
    \begin{tabular}{cccccccc}
        \hline
        \multirow{2}{*}{Size} & \multirow{2}{*}{Method} & \multicolumn{3}{c}{Split CIFAR100} & \multicolumn{3}{c}{Split TinyImageNet} \\ \cline{3-8} 
        &  & MAA & AA & BWT & MAA & AA & BWT \\ \hline \hline
        \multirow{2}{*}{0} & Simsiam & 75.83 & 75.36 & -7.62 & 62.10 & 62.98 & -6.56 \\
        & Barlow-twins & 74.81 & 75.03 & -8.19 & 61.70 & 61.04 & -9.13 \\ \hline
        \multirow{2}{*}{20} & Simsiam & 76.07 & 75.81 & -6.26 & \textbf{63.23} & 64.42 & -5.18 \\
        & Barlow-twins & 75.29 & 75.18 & -8.82 & 63.12 & 63.58 & -7.27 \\ \hline
        \multirow{2}{*}{40} & Simsiam & \textbf{76.37} & 76.20 & -5.97 & 62.70 & 64.42 & -4.56 \\
        & Barlow-twins & \textbf{76.20} & 76.77 & -6.11 & \textbf{63.82} & 64.22 & -6.36 \\ \hline
\end{tabular}}
\end{table}

\paragraph{Analysis of Different Codebook setting}
To further validate the effectiveness of local patterns in CUCL, we conduct experiments using different quantizers. We exclude the interference of rehearsal by conducting the experiments under no rehearsal samples. 
The experimental settings involve fixing the dimension of the representation to 16. 
To maintain the information contained in the quantizer equally, the number of codewords varies according to the number of codebooks.
Additionally, the memory usage is positively correlated with the number of codewords, which is exponentially related to the bits. 
To keep the cost manageable, we apply the 16-bit quantization in this ablation study. 
The experimental results, shown in Tab. \ref{tab:codebook_results}, indicate that using only one codebook results in low performance due to the absence of local patterns. 
Moreover, as the number of codebooks increasing, the different codewords contribute distinct information to the quantized representation, resulting in more diverse and informative representations.
These representations contribute to better final performances. 
However, the results of 8 codebooks on TinyImageNet degraded significantly. 
We posit that the reason is the lack of enough codewords for this harder task, and the reorganization of the quantized subvectors mislead the learning of the model. 
Comparing the results with those in Tab. \ref{tab:ablation_rehearsal} (61.33\% \textit{vs.} 62.10\%), increasing the number of codewords to 8 can capture the local patterns well and correct the learning process.

\paragraph{Analysis of Different Memory Size}
In this ablation study, to explore the effectiveness of our rehearsal algorithm, we conduct experiments with different settings of $S$, including 0, 20, and 40. 
The quantizer in CUCL consist of 8 codebooks, each of which has 8 codewords. 
As shown in Tab. \ref{tab:ablation_rehearsal} on the CIFAR100 dataset, the rehearsal samples not only alleviate catastrophic forgetting but are also beneficial for unsupervised learning. 
Because, unsupervised learning approaches learn discriminative features rather than class-specific information which is learned by supervised methods. 
The more different class samples, the more information about classification boundaries is learned.
Additionally, it is noteworthy that when Barlow-twins saves 20 samples for rehearsal on the CIFAR100 dataset, the BWT behaves worse as MAA and AA increase. 
By checking the results of every task after training on each task, we find that the rehearsal samples promote plasticity. 
However, the stability do not increase comparably to plasticity, resulting in a worse BWT result. 
Meanwhile, the balance between plasticity and stability of Barlow-twins worked well on the TinyImageNet dataset as the size increased. 
Furthermore, on the TinyImageNet dataset, we can observe that the results of Simsiam with 20 samples are worse than those with 40 samples in terms of AA and BWT.
However, comparing the specific results, we find that the first is better than the latter in deed.
This is consistent with the results in term of MAA metric.
We argue that a possible reason is also the imbalance leading to fluctuations in learning.
Overall, these results demonstrate the benefits of our rehearsal algorithm.

\begin{table}
\caption{Experiment results on CIFAR100 dataset with different training time.
The best results are highlighted in bold.}
\label{tab:epoch_results}
\renewcommand\arraystretch{1.3}
\renewcommand\tabcolsep{4.0pt}
\centering
\begin{tabular}{cccccccc}
    \hline
    \multirow{2}{*}{Epoch} & \multirow{2}{*}{CUCL} & \multicolumn{3}{c}{Simsiam} & \multicolumn{3}{c}{Barlow-twins} \\ \cline{3-8}
    & & MAA & AA & BWT & MAA & AA & BWT \\ \hline \hline
    
    \multirow{2}{*}{200} & \textit{w}/\textit{o} & 67.48 & 73.87 & -1.20 & 72.91 & 72.96 & -8.47 \\ 
    & \CC \textit{w}/ & \CC 76.07 & \CC 75.81 & \CC -6.26 & \CC 75.29 & \CC 75.18 & \CC -8.82 \\ \cdashline{0-7}
    
    \multirow{2}{*}{300} & \textit{w}/\textit{o} & 71.75 & 74.95 & -4.63 & 74.58 & 75.07 & -7.66 \\ 
    & \CC \textit{w}/ & \CC 77.84 & \CC 77.15 & \CC -6.87 & \CC 76.40 & \CC 74.91 & \CC -10.27 \\ \cdashline{0-7}
    
    \multirow{2}{*}{400} & \textit{w}/\textit{o} & 74.77 & 75.05 & -8.07 & 75.40 & 75.49 & -7.57 \\ 
    & \CC \textit{w}/ & \CC \textbf{78.34} & \CC 75.45 & \CC -9.26 & \CC \textbf{76.45} & \CC 75.02 & \CC -10.41\\ \cdashline{0-7}
\vspace{-2em}
\end{tabular}
\end{table}

\begin{figure}[t]
\centering
\includegraphics[width=1.0\linewidth]{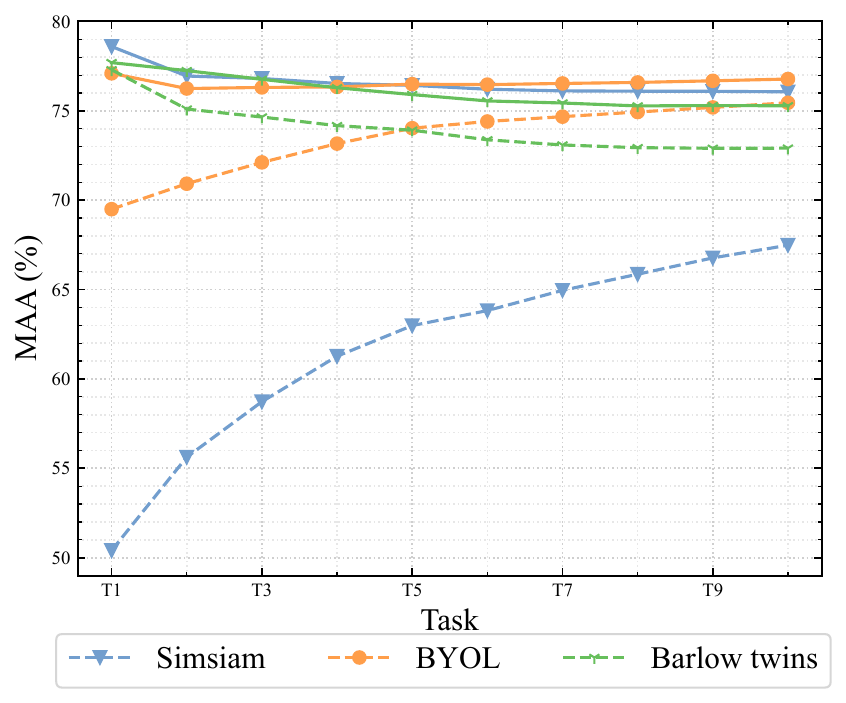}
\vspace{-2em}
\caption{The learning curves in terms of MAA on the Split CIFAR100.
The dashed lines shows the results of finetuning these methodologies.
The solid lines are the results when they are incorporated with our approach.}
\vspace{-2em}
\label{fig:MAA}
\end{figure}

\paragraph{Analysis of Longer Training}
\label{sec:traingtime}
To validate our second assumption in Sec. \ref{sec:intro} that one reason for the poor performance of the initial few tasks is due to slow convergence, we conduct experiments with different numbers of epochs on the CIFAR100 dataset, including  200, 300, and 400 epochs.
The outcomes are shown in Table \ref{tab:epoch_results}.
We observe that enhancing the number of training epochs resulting in better performance for Simsiam and Barlow-twins.
However, the results of Simsiam still suffer from suboptimal issue, which demonstrates that this assumption \ref{sec:intro} is false.
In addition, the results of 400-epoch Simsiam training are still inferior to those of 200-epoch Simsiam with CUCL training.

\subsubsection{Analysis of the training process}
\label{MAATracking}
To further evaluate the learning ability of model during the training process, we plot the results on CIFAR100 in terms of MAA metric. 
Three models including Simsiam, BYOL, and Barlow-twins are considered as unsupervised backbones.
Dashed lines represent results when the models are fine-tuned, while solid curves show the performance when our CUCL is integrated.
The final learning curves are shown in Fig. \ref{fig:MAA}.
These curves reveal that Simsiam and BYOL face the suboptimal issue. 
Specifically, we observe considerate improvements on Simsiam on the first (50.4\% \textit{vs}. 78.6\%), and on the end (67.48\% \textit{vs}. 76.07\%).
Furthermore, Barlow-twins exhibits good performance in the initial task, but it suffers catastrophic forgetting (77.3\% \textit{vs}. 72.91\%).
Nonetheless, our approach enhances the first task's outcome to 77.7\%. 
In addition, the effective rehearsal method helps Barlow-twins to retain previous acquired knowledge, leading to a 75.29\% outcome.

\section{Conclusion}
This paper investigated the unsupervised continual learning paradigm.
Our results revealed a problem that the performances of the first few tasks are suboptimal.
% Conventional continual learning metrics cannot capture this issue.
% Therefore, we proposed Mean Average Accuracy (MAA), a new metric that can measure the performance of each task rather than only the last one.
Additionally, to ameliorate this problem, we proposed Codebook for
Unsupervised Continual Learning (CUCL) to confer diversity for representations and apply the contrastive learning on the original representation and the quantized one from a different view to guide the model to capture discriminative features.
The outcomes of extensive experiments demonstrated the efficacy of our proposed method.

\begin{acks}
This study is supported by grants from National Key R\&D Program of China (2022YFC2009903/2022YFC2009900), the National Natural Science Foundation of China (Grant No. 62122018, No. 62020106008, No. 61772116, No. 61872064), Fok Ying-Tong Education Foundation(171106), and SongShan Laboratory YYJC012022019. 
\end{acks}


\balance
\bibliographystyle{ACM-Reference-Format}
\bibliography{ref}

\end{document}